%% file: main.tex
\documentclass[11pt]{article}
\usepackage{acl2016}
\usepackage{times}
\usepackage{url}
\usepackage{bm}
\usepackage{latexsym}
\usepackage{graphicx}
\usepackage[usenames,dvipsnames,svgnames,table]{xcolor}
\usepackage{gb4e}
\noautomath
\usepackage{varwidth}
\usepackage{tikz}
\usetikzlibrary{calc, backgrounds, trees, fit, positioning, arrows, chains, shapes.geometric, decorations.pathreplacing, decorations.pathmorphing,shapes, matrix, shapes.symbols}
\usepackage{tikz-dependency}
\noautomath
\usepackage{fixltx2e}
\usepackage{csquotes}
\usepackage{booktabs}
\usepackage{multirow}
\usepackage[table]{xcolor}
\usepackage{subcaption}
\aclfinalcopy % Uncomment this line for the final submission
\def\aclpaperid{26} %  Enter the acl Paper ID here

\newcolumntype{x}[1]{>{\centering\arraybackslash\hspace{0pt}}p{#1}}
\definecolor{tbl}{rgb}{.85,.95,.85}

\title{This before That:\\Causal Precedence in the Biomedical Domain}

\author{Gus Hahn-Powell\quad
Dane Bell\quad
Marco A. Valenzuela-Esc\'{a}rcega\quad
Mihai Surdeanu \\ % supervisor goes last
\\University of Arizona \\ 
Tucson, AZ 85721, USA \\
{\tt hahnpowell@email.arizona.edu} \\
}
\date{}

\begin{document}
\maketitle
\begin{abstract}
Causal precedence between biochemical interactions is crucial in the biomedical domain, because it transforms collections of individual interactions, e.g., bindings and phosphorylations, into the causal mechanisms needed to inform meaningful search and inference. Here, we analyze \textit{causal} precedence in the biomedical domain as distinct from open-domain, \textit{temporal} precedence. 
First, we describe a novel, hand-annotated text corpus of causal precedence in the biomedical domain. Second, we use this corpus to investigate a battery of models of precedence, covering rule-based, feature-based, and latent representation models.
The highest-performing individual model achieved a micro F1 of 43 points, approaching the best performers on the simpler temporal-only precedence tasks. Feature-based and latent representation models each outperform the rule-based models, but their performance is complementary to one another. We apply a sieve-based architecture to capitalize on this lack of overlap, achieving a micro F1 score of 46 points.
\end{abstract}

\input{intro}

\input{corpus}

\input{models}

\input{results}

\input{discussion}
\input{related}

\section{Conclusion}

These are the first experiments regarding automatic annotation of causal precedence in the biomedical domain. Although the 
dearth of temporal expressions and other regular linguistic cues make the task especially difficult in this domain, the initial 
results are promising, and demonstrate that a sieve-based system of the models tested here improves performance over the top-performing individual component. 
Both the annotation corpus and the models described here represent large steps toward linking automatic reading to a larger, 
more informative biological mechanism.

%\todo{Add 1 para here for true conclusions. Summarize the work again.}

\section*{Acknowledgments}

This work was funded by the Defense Advanced Research Projects Agency (DARPA) Big Mechanism program under ARO contract W911NF-14-1-0395.

%The acknowledgments should go immediately before the references.  Do
%not number the acknowledgments section. Do not include this section
%when submitting your paper for review.
%\cite{chambers2014dense} % (author, year)
%\shortcite{APA:83} % (year)
%\newcite{APA:83} % author name (year)

%\newpage

\bibliography{sieves}
\bibliographystyle{acl2016}

%\appendix
%
%\section{Supplemental Material}
%\label{sec:supplemental}
%ACL 2016 also encourages the submission of supplementary material
%to report preprocessing decisions, model parameters, and other details
%necessary for the replication of the experiments reported in the 
%paper. Seemingly small preprocessing decisions can sometimes make
%a large difference in performance, so it is crucial to record such
%decisions to precisely characterize state-of-the-art methods.

\end{document}

%% file: intro.tex
\section{Introduction}
 % Deadline: May 8 (11:59 PM Eastern US)
 % submit via https://www.softconf.com/acl2016/BioNLP16/
 % workshop: Berlin, Germany, August 12 -13, 2016
 % Short papers may consist of up to four (4) pages of content, plus unlimited references. Appropriate short paper topics include preliminary results, application notes, descriptions of work in progress, etc.
 % Full papers should not exceed eight (8) pages of text, plus unlimited references. These are intended to be reports of original research. BioNLP aims to be the forum for interesting, innovative, and promising work involving biomedicine and language technology, whether or not yielding high performance at the moment. This by no means precludes our interest in and preference for mature results, strong performance, and thorough evaluation. Both types of research and combinations thereof are encouraged.

In the biomedical domain, an enormous amount of information about protein, 
gene, and drug interactions appears in the form of natural language across 
millions of academic papers. There is a tremendous ongoing effort 
\cite{nedellec2013overview,kim2012genia,kim2009overview} to extract individual chemical 
interactions from these texts, but these interactions are only isolated fragments of 
larger causal mechanisms such as protein signaling pathways. Nowhere, however, including any database, is the complete 
mechanism described in a form that lends itself to causal search or inference. 
The absence of such a database is not for lack of trying; Pathway Commons 
\cite{cerami2011pathway} aims to address the need, but its authors estimate 
it currently covers 1\% of the literature due to the high cost of annotation\footnote{Personal communication.}.
This issue only grows more pressing with the yearly growth in biomedical 
publishing, which presents an otherwise insurmountable challenge for 
biomedical researchers to query and interpret.

The Big Mechanism program~\cite{cohen2015} aims to construct 
exactly such large-scale mechanistic information by reading and assembling protein signaling pathways that are relevant for cancer, and exploit them to generate novel explanatory and treatment hypotheses.
Although prior work \cite{chambers2014dense,1604.08120} has addressed the 
challenging area of temporal precedence in the open domain, the biomedical 
domain presents very different data and, consequently, requires novel 
techniques. Precedence in mechanistic biology is 
\textit{causal} rather than \textit{temporal}.  Though event temporality is crucial to understanding electronic health records for individual patients \cite{bethard2015semeval,bethard2016semeval}, 
its contribution to the understanding of biomolecular reactions is less clear as these events and processes may repeat in extremely short cycles, continue without end, or overlap in time.  
At any level of abstraction, causal precedence encodes mechanistic information and facilitates inference over spotty 
evidence. For the purpose of this work, \textit{precedence} is defined for two events, A and B, as
\begin{quote}
A precedes B if and only if the output of A is necessary for the successful execution of B.\footnote{See the ``precedes'' examples in Table~\ref{tab:label-examples}.}
\end{quote}

Very little annotated data exists for causal precedence, especially efforts focusing on signaling pathways.  BioCause~\cite{Mihăilă2013}, for instance, is centered on connections between claims and evidence and contains only 51 annotated examples of causal precedence\footnote{These are marked in the BioCause corpus as \texttt{Causality} events with \texttt{Cause} and \texttt{Effect} arguments. The remaining 800 annotations are claim-evidence relations.}.  Our work\footnote{The corpus, tools, and system introduced in this work are publicly available at \url{https://github.com/myedibleenso/this-before-that}} offers three contributions in aid of automatically extracting 
causal ordering in biomedical text. First, we provide and describe a 
dataset of real text examples, manually annotated for causal precedence. 
Second, we analyze the efficacy of a battery of different models in automatically determining precedence, built on top of 
the Reach automatic reading system~\cite{reach2015,Valenzuela+:2015aa} and measured against 
this novel corpus. In particular, we investigate three classes of models: (a) deterministic rule-based models inspired by the precedence sieves proposed by~\newcite{chambers2014dense}, (b) feature-based models, and (c) models that rely on latent representations such as long short-term memory (LSTM) networks~\cite{hochreiter1997long}. Our analysis indicates that while independently the top-performing model achieves a micro F1 of 43, these models are largely complementary with a combined recall of 58 points.
Lastly, we conduct an error analysis of these models to 
motivate and inform future research.

%% file: corpus.tex
\section{A Corpus for Causal Precedence in the Biomedical Domain}

\begin{table}[!tb]
\centering
\small
\begin{tabular}{l p{4.75cm}}
\toprule
\multicolumn{1}{c}{\textit{Relation}} & \multicolumn{1}{c}{\textit{Example}} \\
\midrule
% causal precedence
\textbf{E1 precedes E2} & {A is phosphorylated by B.  \newline
Following its phosphorylation, A binds with C.} \\   
\midrule
\textbf{E2 precedes E1} &
{A is phosphorylated by B. \newline
Prior to its phosphorylation, A binds with D.} \\
\midrule
% equivalence
\textbf{Equivalent} & {
The phosphorylation of A by B. \newline
A is phosphorylated by B.} \\
\midrule
% subsumption
\textbf{E1 specifies E2} & {A is phosphorylated by B at Site 123. \newline 
A is phosphorylated by B.} \\
\midrule
\textbf{E2 specifies E1} & {
A is phosphorylated by B. \newline
A is phosphorylated by B at Site 123.} \\
\midrule
% Other
\textbf{Other} & {
B does not regulate C when C is bound to A.} \\
\midrule
% None
\textbf{None} & {
A phosphorylates B. \newline
A ubiquitinates C.} \\
\bottomrule
\end{tabular}
\caption{The seven inter-event relation labels annotated in the corpus.  The ``precedes'' labels are causal.  Subsumption is captured with the ``specifies'' labels.}
\vspace{-4mm}
\label{tab:label-examples}
\end{table}

% add figures of annotation proportions in intra- and intersentence relations
% examine distribution of sentence lengths with each type -- Dane's suspicion is that longer sentences, having more errorful parses, are also harder to extract events from, but overall should have more precedence relations intra sentence.
% review process of annotation; explain ambiguity with a couple of examples -- is "A upregulates the B-induced phosphorylation of C" a regulation of a regulation?; mention Dane's checks against diagrams unavailable to text reader?
Our corpus annotates several types of relations between mentions of biochemical interactions. Following common terminology promoted by the BioNLP shared tasks, we will interchangeably use ``events'' to refer to these interactions.
To generate candidate events for our planned annotations, we ran the Reach event extraction system~\cite{reach2015,Valenzuela+:2015aa} over the full text\footnote{We chose to ignore the ``references'', ``materials'', and 
``methods'' sections, which generally do not contain mechanistic information.} of 500 biomedical papers taken from the Open 
Access subset of PubMed\footnote{\url{http://www.ncbi.nlm.nih.gov/pmc/tools/openftlist/}}. The events extracted by Reach are 
biochemical events of two types: simple events such as phosphorylation that modify one or more entities (typically proteins), and 
nested events (regulations) that have other events as arguments.

To improve the likelihood of finding pairs of events with a relevant link, we filtered event pairs by imposing the following requirements for inclusion in the corpus:

\begin{enumerate}
 \item \textit{Event pairs must share at least one participant.} This constraint is based on the observation that interactions that share participants are more likely to be connected.
 \item \textit{Event pairs must be within 1 sentence of each other.} Similarly, discourse proximity increases the likelihood of two events being related.
 \item \textit{Event pairs must not share the same type.}  This helps to maximize the diversity of the dataset.
 \item \textit{Event pairs must not already be contained in an extracted Regulation event.} For example, we did not annotate the relation between the binding and the phosphorylation events in ``The binding of X and Y is inhibited by X phosphorylation'', because it is already captured by most state-of-the-art biomedical event extraction systems. 
\end{enumerate}

After applying these constraints, only 1700 event pairs remained.  In order to rapidly annotate the event pairs, we developed a browser-based annotation UI that is completely client-side (see Figure~\ref{fig:annotation-tool}).  Using this tool, we annotated 1000 event pairs for this work; 84 of these were discarded due to severe extraction errors.  The annotations include the event spans, event triggers (i.e., the verbal or nominal predicates that indicate the type of interaction such as ``binding'' or ``phosphorylated''), source document, minimal sentential span encompassing both event mentions, and whether or not the event pair involves coreference for either the event trigger or the event participants.  For events requiring coreference resolution, we expanded the encompassing span of text to also capture the antecedent.  Note that domain-specific coreference resolution is a component of the event extraction system used here~\cite{1603.03758}. 

When describing the relations between these event pairs, we refer to the event that occurs first in text as Event 1 (E1) and the event that follows as Event 2 (E2).  Each (E1, E2) pair was assigned one of seven labels: ``E1 precedes E2'', ``E2 precedes E1'', ``Equivalent'', ``E1 specifies E2'', ``E2 specifies E1'', ``Other'', or ``None''.  Table~\ref{tab:label-examples} provides examples for each of these labels. 
We converged on these labels because they are fundamental to the assembly of causal mechanisms from a collection of events. Collectively, the seven labels address three important assembly tasks: {\em equivalence}, i.e., understanding that two event mentions discuss the same event, {\em subsumption}, i.e., the two mentions discuss the same event, but one is more specific than the other, and, most importantly, {\em causal precedence}, the identification of which is the focus of this work.  During the annotation process, we came across examples of other relevant phenomena.  We grouped these instances under the label ``Other'' and leave their analysis for future work.

Though simplified, the examples in Table~\ref{tab:features} illustrate that this is a complex task sensitive to linguistic evidence. For example, the direction of the precedence relations in the first two rows in the table changes based on a single word in the context (``prior'' vs. ``following'').

\begin{figure}[t]
    \includegraphics[trim=2.15cm 1.95cm 1.75cm 2cm, clip, width=\columnwidth]{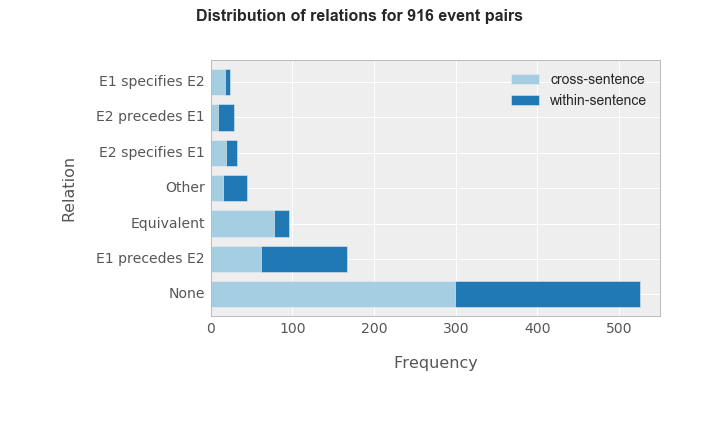}
    \caption{The distribution of assembly relation labels both within and across sentences.}  
    \vspace{-4mm}
    \label{fig:relation-label-dist}
\end{figure}

\begin{figure}[t]
    \includegraphics[trim=2.15cm 1.95cm 1.75cm 2cm, clip, width=\columnwidth]{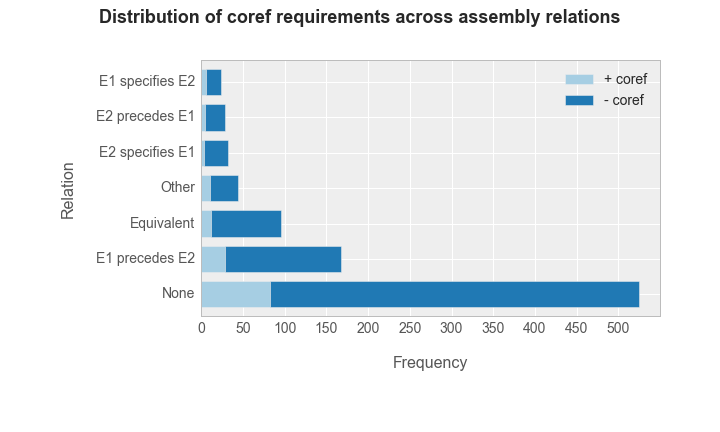}
    \caption{The distribution of event pairs involving coreference across assembly relations.}  
    \vspace{-4mm}
    \label{fig:relation-coref-dist}
\end{figure}

In terms of the distribution of relations, causal precedence pairs appear more frequently within the same sentence, while cases of the subsumption (``specifies'') and equivalence relations are far more common across sentences (see Figure~\ref{fig:relation-label-dist}).  Coreference is involved in 10--15\% of the instances for each relation label (see Figure~\ref{fig:relation-coref-dist}). 

\begin{figure*}[tb]
\centering
\includegraphics[scale=0.8, trim={0 8.4cm 0 0.5cm}, clip, width=\textwidth]{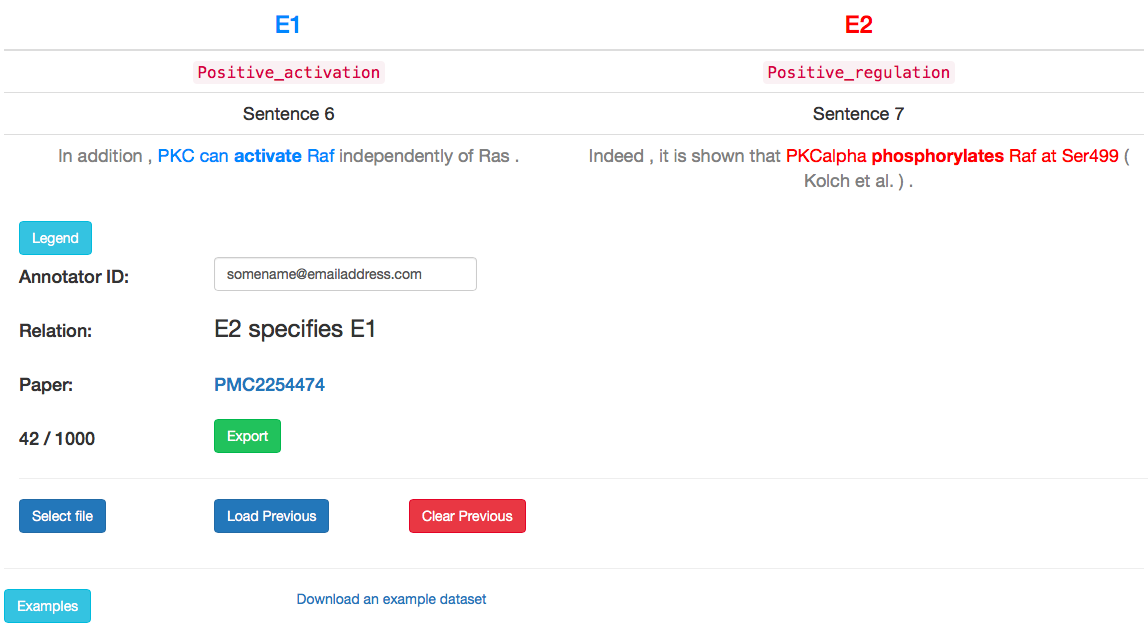}
\caption{Browser-based tool for annotating assembly relations in text.  An annotation instance consists of a pair of event mentions.  The annotator assigns a label to each pair of events using the number keys and navigates from annotation to annotation using the arrow keys.  E1 refers to the event in the pair that appears first in the text.  The event span is formatted to stand out from the surrounding text.  The ``Paper'' field provides the annotator with easy access to the full text of the source document for the current annotation instance.  Annotations can be exported to JSON and reloaded via a local storage cache or through file upload.}\label{fig:annotation-tool}
\label{fig:annotation-tool}
\end{figure*}
The annotation process was performed by two linguists familiar with the biomedical domain. To minimize errors, the annotation task was initially performed together at the same workstation.\footnote{Similar to pair programming.}  On a randomly selected sample of 100 event pairs, the two annotators had a Cohen's kappa score~\cite{cohen1960coefficient} of 0.82, indicating ``almost perfect'' agreement for the \textit{precedes} labels \cite{landis1977measurement}.

%% file: models.tex
%%%%%%%%%%%%%%%%%
% Models
%%%%%%%%%%%%%%%%%
\section{Models of Causal Precedence}

We have developed both deterministic, interpretable models and automatic, machine-learning models for detecting causal precedence in our dataset. 
Importantly, the models covered in this work focus solely on causal precedence, which is the most complex relation annotated in the dataset previously introduced. Thus, for all experiments discussed here, we reduce these annotations to three labels: ``E1 precedes E2'', ``E2 precedes E1'', and {\tt Nil}, which covers all the other labels in the corpus.

\begin{table}[h]
%\small
\centering
\begin{tabular}{l l}
\toprule
\textit{Model}	& \textit{Rules}	\\
\midrule
Intra-sentence	& 29	\\
Inter-sentence	& 5	\\
Reichenbach		& 8	\\
\bottomrule
\end{tabular}
\caption{Few rules defined each deterministic model of precedence compared with the number of features for the machine learning models.}\label{tab:rules}
\vspace{-4mm}
\end{table}

\input{deterministic}
\input{features}
\input{latent}

%% file: deterministic.tex
% causal ordering
\subsection{Deterministic Models}

The deterministic models are defined by a small number of hand-written rules using the Odin event extraction framework \cite{1509.07513}. The number of rules for each model is shown in Table \ref{tab:rules}, and sharply contrast with the 92,711 features introduced later (Table \ref{tab:features}) that are used by our machine-learning models. In order to avoid overfitting, all of the deterministic models were created without reference to the annotation corpus, using general linguistic expertise and domain knowledge.

\paragraph{Intra-sentence ordering}

Within sentences, syntactic regularities can be exploited to cover a large variety of grammatical constructions indicating precedence relations. Rules defined over dependency parses~\cite{de2008stanford} capture precedence in sentences like those in (\ref{ex:intra1}) and (\ref{ex:intra2}) as well as many others.

\begin{exe}
	\ex\label{ex:intra1} $[$The RBD of PI3KC2B binds HRAS$]$\textsubscript{after} , when $[$HRAS is not bound to GTP$]$\textsubscript{before}
	\ex\label{ex:intra2} $[$The ubiquitination of A$]$\textsubscript{before} is followed by $[$the phosphorylation of B$]$\textsubscript{after}
\end{exe}
Other phrases captured include: ``precedes'', ``due to'', ``leads to'', ``results in'', etc.

\paragraph{Inter-sentence ordering}

Although syntax operates over single sentences, cross-sentence time expressions can indicate ordering, as shown in Examples (\ref{ex:inter1}) and (\ref{ex:inter2}). We exploit these regularities as well by checking for sentence-initial word combinations.

%\begin{exe}
%	\ex $[$A is phosphorylated by B$]$\textsubscript{before}.  As a down\-stream effect, $[$C is \textellipsis$]$\textsubscript{after}
%	\ex $[$A is phosphorylated by B$]$\textsubscript{before}.  Later, $[$C is \textellipsis$]$\textsubscript{after}
%	\ex $[$A is phosphorylated by B$]$\textsubscript{before}.  In response, $[$C is \textellipsis$]$\textsubscript{after}
%	\ex $[$A is phosphorylated by B$]$\textsubscript{before}.  Ultimately, $[$C is \textellipsis$]$\textsubscript{after}
%	\ex $[$A is phosphorylated by B$]$\textsubscript{after}.  For this, $[$A is \textellipsis$]$\textsubscript{before}
%\end{exe}

\begin{exe}
	\ex\label{ex:inter1} $[$A is phosphorylated by B$]$\textsubscript{before}.  As a down\-stream effect, $[$C is \textellipsis$]$\textsubscript{after}
	\ex\label{ex:inter2} $[$A is phosphorylated by B$]$\textsubscript{before}.  $[$C is then \textellipsis$]$\textsubscript{after}
\end{exe}
Other phrases captured include: ``Later'', ``In response'', ``For this'', and ``Ultimately''.

\paragraph{Verbal tense- and aspect-based (Reichenbach) ordering}

Following \newcite{chambers2014dense}, we use deterministic rules to establish precedence between events that have certain verbal tense and aspect. These rules are derived from linguistic analysis of tense and aspect by \cite{reichenbach1947,derczynski2013empirical}. Example (\ref{ex:reichenbach}) illustrates a case in which we can accurately infer order just from this information. Because \textit{has been phosphorylated} has past tense and perfective aspect, this model concludes that it precedes \textit{share} (present tense, simple aspect) and thus the binding of histone H2A.

\begin{exe}
	\ex\label{ex:reichenbach} These [PTIP] proteins also share the ability to bind histone H2A (or H2AX in mammals) that has been phosphorylated\textellipsis.
\end{exe}

The logic determining which tense-aspect combinations receive which precedence relations is identical to CAEVO, which is possible because it is open source\footnote{\url{https://github.com/nchambers/caevo}}. However, CAEVO operates over annotations that include gold tense and aspect values, whereas this model additionally detects tense and aspect using Odin rules before applying this logic.

%% file: features.tex
\subsection{Feature-based Models}

\begin{table*}[!htb]
\centering
\scriptsize
\begin{tabular*}{\textwidth}{p{1.5cm} p{2cm} p{11.25cm}}
\toprule
& \textit{Feature} & \textit{Description} \\
\midrule
% Event features
\multirow{12}*{\parbox{1.5cm}{\centering \scriptsize{\textit{Event}}}}& Event labels & The taxonomic labels Reach assigned to the event (e.g.\ \textit{phosphorylation} $\rightarrow$ Phosphorylation, AdditiveEvent, \textellipsis). \\
\cmidrule{2-3}
& Event trigger & The predicate signaling an event mention (ex. ``phosphorylated'', ``phosphorylation''). \\
\cmidrule{2-3}
& {Event trigger + label} & A concatenation of the event's trigger with the event's label. \\
\cmidrule{2-3}
& token \textit{n}-grams with entity replacement & \textit{n}-grams of the tokens in the mention span, where each entity is replaced with the entity label (ex. ``the ABC protein'' $\rightarrow$ ``the PROTEIN'').  If an entity is shared between pairs of events, replace it with the label SHARED. \\
\cmidrule{2-3}
& token \textit{n}-grams with role replacement & \textit{n}-grams of the tokens in the mention span, where each argument is replaced with the argument role (ex. ``A inhibits the phosphorylation of B'' $\rightarrow$ ``CONTROLLER inhibits the CONTROLLED'') \\
\cmidrule{2-3}
 & {Syntactic path from \newline trigger to args} & Variations of the syntactic dependency path from an event's trigger to each of its arguments (unlexicalized path, path + lemmas, trigger $\rightarrow$ argument role, trigger $\rightarrow$ argument label, etc.). \\
\midrule
% event-event (surface)
\multirow{1}{*}{\parbox{1.5cm}{\centering\scriptsize{\textit{Event-Event} \newline (surface)}}}  & Interceding tokens (\textit{n}-grams) & \textit{n}-grams (1-3) of the tokens between E1 and E2. \\
\midrule
% event-event (syntax)
\multirow{8}[8]*{\parbox{1.5cm}{\centering\scriptsize{\textit{Event-Event} \newline (syntax)}}} & {Cross-sentence \newline syntactic paths} & A concatenation of the syntactic path from the sentential \textsc{ROOT} to an event's trigger (see the example in Figure~\ref{fig:cross-sentence-syntax}). \\
\cmidrule{2-3}
& {Trigger-to-trigger syntactic paths \newline (within sentence)} & the syntactic path from the trigger of E1 to the trigger of E2 \\
\cmidrule{2-3}
& Shortest syntactic paths & The shortest syntactic path between E1 and E2 (restricted to intra-sentence cases). \\
\cmidrule{2-3}
& Syntactic distance & The length of each syntactic path (restricted to intra-sentence cases). \\
\midrule
% Coref features
\multirow{6}[10]*{\parbox{1.5cm}{\centering \scriptsize{\textit{Coreference}}}} & \textit{Event} features for anaphors & Whether or not an event mention is resolved through coreference.  For cases of coreference, generate the \textit{Event} features prefixed with ``coref-anaphor'' for the text labeled ``E1-anaphor'' in the following example:  
\begin{exe}
\ex $[$A binds with B$]$\textsubscript{E1-antecedent}
\ex $[$This interaction$]$\textsubscript{E1-anaphor} precedes the $[$phosphorylation of C$]$\textsubscript{E2}   
\end{exe} \\
\cmidrule{2-3}
& Resolved arguments & Which arguments, if any, were resolved through coreference. For example: 

$[$\underline{The mutant}\textsubscript{\textsc{theme}} binds with B\textsubscript{\textsc{theme}}$]$\textsubscript{E1} $\rightarrow$ \texttt{\textsc{theme}:resolved} \\
\bottomrule
\end{tabular*}
    
\caption{An overview of the primary features used in the feature-based classifier, grouped into four classes: {\em Event} -- features extracted from the two participating events, in isolation; {\em Event-Event (surface)} -- features that model the lexical context between the two events; {\em Event-Event (syntax)} -- features that model the syntactic context between the two events; and {\em Coreference} -- features that capture coreference resolution information that impact the participating events.}
\label{tab:features}
\end{table*}

Most instances of causal precedence cannot be captured with deterministic rules, because they lack explicit words, phrases, or syntactic structures that unambiguously mark the relation. Using a combination of the surface, syntactic, and taxonomic features outlined in Table~\ref{tab:features}, we trained a set of statistical classifiers to detect causal precedence relations between pairs of events in our corpus.  For training and testing purposes, we treated any instance not labeled as either ``E1 precedes E2'' or ``E2 precedes E1'' as a negative example.  
We examined the following statistical models: a linear kernel SVM~\cite{CC01a}, logistic regression~\cite{Fan:2008:LLL:1390681.1442794}, and random forest\footnote{Abbreviated as RF}~\cite{processors2014}.  For the SVM and logistic regression (LR) models, we also compared the effects of L1 and L2 regularization.  
% ms: too soon to mention results.
%See Table~\ref{tab:prec-results} for a summary of the results.

% cross-sentence syn path fig
\begin{figure*}[!tbh]
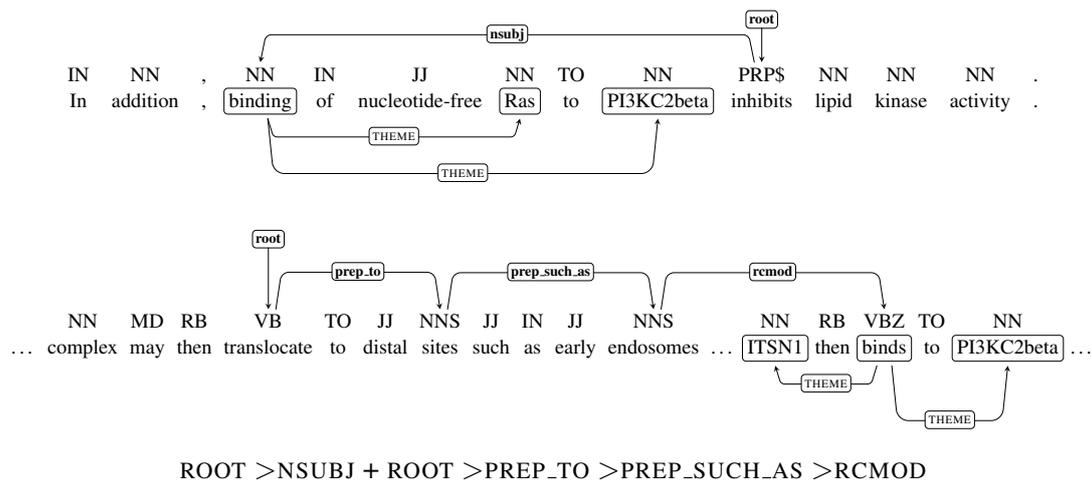

\begin{displayquote}
\small
In addition, \textbf{binding of nucleotide-free Ras to PI3KC2\bm{$\beta$}} inhibits its lipid kinase activity.  The PI3KC2$\beta$ and Ras complex may then translocate to distal sites such as early endosomes (EE) where \textbf{ITSN1 then binds to PI3KC2\bm{$\beta$}} leading to the release of nucleotide-free Ras and activation of the lipid kinase activity of PI3KC2$\beta$.
\end{displayquote}

% sentence 1 (E1)
\begin{center}
\scalebox{0.7}{
\begin{dependency}
   \begin{deptext}[column sep=.25cm, row sep=.1ex]
      IN \& NN \& , \& NN \& IN \& JJ \& NN \& TO \& NN \& PRP\$ \& NN \& NN \& NN \& . \\
      In \& addition \& , \& binding \& of \& nucleotide-free \& Ras \& to \& PI3KC2beta \& inhibits \& lipid \& kinase \& activity \& . \\
   \end{deptext}
   \deproot[edge unit distance=1.5ex]{10}{\textbf{root}}{root}
   \depedge[edge unit distance=0.5ex]{10}{4}{\textbf{nsubj}}{nsubj}
   \wordgroup{2}{4}{4}{trigger}
   \wordgroup{2}{7}{7}{theme1}
   \wordgroup{2}{9}{9}{theme2}
   \groupedge[edge below]{trigger}{theme1}{\textsc{theme}}{2ex}
    \groupedge[edge below]{trigger}{theme2}{\textsc{theme}}{6ex}    
\end{dependency}
}
\end{center}

\vspace{0.25cm}

% sentence 2 (E2)
\begin{center}
\scalebox{0.7}{   
\begin{dependency}
   %\begin{deptext}[font=\scriptsize, column sep=.05cm, row sep=.1ex]
   \begin{deptext}[column sep=.05cm, row sep=.1ex]
% The \& PI3KC2beta \& and \& Ras \& complex \& may \& then \& translocate \& to \& distal \& sites \& such \& as \& early \& endosomes \& ( \& EE \& ) \& where \& ITSN1 \& then \& binds \& to \& PI3KC2beta \& leading \& to \& the \& release \& of \& nucleotide-free \& Ras \& and \& activation \& of \& the \& lipid \& kinase \& activity \& of \& PI3KC2beta \& . \\
{} \& NN \& MD \& RB \& VB \& TO \& JJ \& NNS \& JJ \& IN \& JJ \& NNS \& {} \& NN \& RB \& VBZ \& TO \& NN \& {} \\
\dots \& complex \& may \& then \& translocate \& to \& distal \& sites \& such \& as \& early \& endosomes \& \dots \& ITSN1 \& then \& binds \& to \& PI3KC2beta \& \dots \\
   \end{deptext}
   \deproot[edge unit distance=2.25ex]{5}{\textbf{root}}{root}
   \depedge[edge unit distance=1.35ex]{5}{8}{\textbf{prep\_to}}{prepto}
   \depedge[edge unit distance=1ex]{8}{12}{\textbf{prep\_such\_as}}{prepsuchas}
    \depedge[edge unit distance=1ex]{12}{16}{\textbf{rcmod}}{rcmod}
   \wordgroup{2}{16}{16}{trigger}
   \wordgroup{2}{14}{14}{theme1}
   \wordgroup{2}{18}{18}{theme2}
   \groupedge[edge below]{trigger}{theme1}{\textsc{theme}}{2ex}
    \groupedge[edge below]{trigger}{theme2}{\textsc{theme}}{6ex}    
\end{dependency}
}
\end{center}

% path feature
\begin{center}
\textsc{root} \textsc{$>$nsubj} + \textsc{root} \textsc{$>$prep\_to} \textsc{$>$prep\_such\_as} \textsc{$>$rcmod}  
\end{center}

\caption{Generation procedure for the cross-sentence syntactic path feature.  For each event in a pair, we find the shortest syntactic path originating from the sentential root node leading to a token in the event's trigger.  The two syntactic paths are then joined using the $+$ symbol to form a single feature.}
\label{fig:cross-sentence-syntax}
\end{figure*}

%\todo{Future work: explore discourse for cross-sentence paths.}

%% file: latent.tex
%%%%%%%%%%%%%%%%%%
% Latent features
%%%%%%%%%%%%%%%%%%
\subsection{Latent Representation Models}
Due to the complexity of the task and variety of causal precedence instances encountered during the annotation process, it is unclear whether a linear combination of engineered features is sufficient for broad coverage classification.  For this reason, we introduce a latent feature representation model using an LSTM~\cite{hochreiter1997long,bergstra+al:2010-scipy,chollet2015} to capture underlying semantic features by incorporating long-distance contextual information and selectively persisting memory of previous event pairs to aid in classification.

%used categorical cross-entropy

The basic architecture is shown in Figure~\ref{fig:lstm-architecture}.  The input to this model is the provenance of the relation, i.e., the whole text containing the two events and the text in between. Formally, this is represented as a concatenated sequence of 200 dimensional vectors where each vector in the sequence corresponds to a token in the minimal sentential span encompassing the event pair being classified.  
Intuitively, this LSTM ``reads'' the text from left to right and outputs a classification label from the set of three when done.
We consider two variations of this model: the basic model (LSTM) with the vector weights for each token uninitialized and a second form (LSTM+P) where the vectors are initialized using pre-training.  In the pre-training configuration, the vector weights are initialized using word embeddings generated by a word2vec~\cite{mikolov13,rehurek_lrec} model trained on the full text of over 1 million biomedical papers taken from the Open Access subset of PubMed.  Because the corpus is only 1000 annotations, it was thought that pre-training could improve prediction of causal precedence and guide the model with distributional semantic representations specific to this domain.  

\begin{figure}[!htb]
\centering
\begin{tikzpicture}
    [thick, scale=0.8, transform shape, ->, shorten >=2pt, node distance=2.3cm and 0.25cm]
    \tikzstyle{component} = [%
        align=center,
        rectangle, 
        rounded corners,
        draw=black, very thick,
        execute at begin node={\begin{varwidth}{6em}},
        execute at end node={\end{varwidth}},
        text centered,
    ]
    \tikzstyle{output-layer}= [%
        fill=green!10,
        draw=green, very thick
    ]
    \tikzstyle{shortened}= [yshift=0.8cm]
    % left fork
    \node [component](left-fork){Input 1 \\ \small{E1 tokenized} \\ \small{text}};
    \node [component, shortened][below of = left-fork](left-embeddings){Embeddings};
    \node [component, shortened][below of = left-embeddings](left-lstm){LSTM};
    \node [component, shortened][below of = left-lstm](left-dropout){Dropout};
    % middle fork
    %above right=of
    \node [component][right of = left-fork](middle-fork){Input 2 \\ \small{Encompassing} \\ \small{tokenized text}};
    \node [component, shortened][below of = middle-fork](middle-embeddings){Embeddings};
    \node [component, shortened][below of = middle-embeddings](middle-lstm){LSTM};
    \node [component, shortened][below of = middle-lstm](middle-dropout){Dropout};
    % right fork
    \node [component][right of = middle-fork](right-fork){Input 3 \\ \small{E2 tokenized} \\ \small{text}};
    \node [component, shortened][below of = right-fork](right-embeddings){Embeddings};
    \node [component, shortened][below of = right-embeddings](right-lstm){LSTM};
    \node [component, shortened][below of = right-lstm](right-dropout){Dropout};
    
    % merge
    \node [component][below of = middle-dropout, yshift=0.5cm](merge){Merge \\ \small{(concat)}};
    \node [component, shortened][below of = merge](merge-dropout){Dropout};
    \node [component, shortened][below of = merge-dropout](merge-dense){Dense};
    \node [component, output-layer, shortened][below of = merge-dense](merge-softmax){Softmax};
    
    % edges
    \path (left-fork) edge (left-embeddings);
    \path (left-embeddings) edge (left-lstm);
    \path (left-lstm) edge (left-dropout);

    \path (middle-fork) edge (middle-embeddings);
    \path (middle-embeddings) edge (middle-lstm);
    \path (middle-lstm) edge (middle-dropout);
    
    \path (right-fork) edge (right-embeddings);
    \path (right-embeddings) edge (right-lstm);
    \path (right-lstm) edge (right-dropout);
    
    \path (left-dropout) edge (merge);
    \path (middle-dropout) edge (merge);
    \path (right-dropout) edge (merge);
    
    \path (merge) edge (merge-dropout);  
    \path (merge-dropout) edge (merge-dense);
    \path (merge-dense) edge (merge-softmax);     

\end{tikzpicture}
\caption{Modified architecture for a latent feature model with three-pronged input: the text of event 1 (left), the minimal sentential span encompassing events 1 and 2 (middle), and the text of event 2 (right).}
\label{fig:pitchfork-lstm-architecture}
\end{figure}
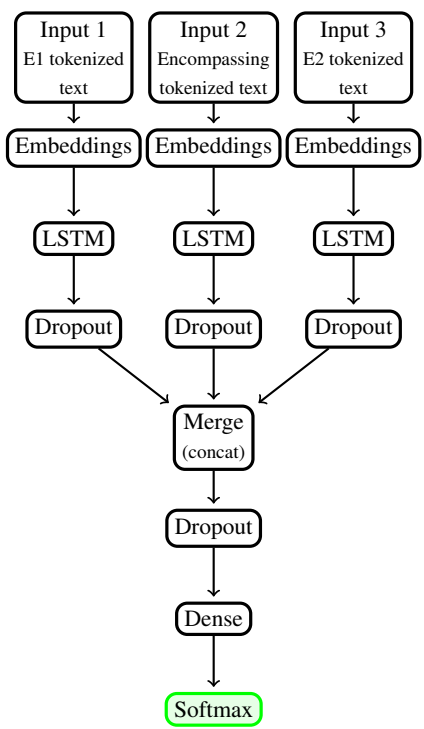

Building on this simple blueprint, we designed a three-pronged ``pitchfork'' (FLSTM) where the span of E1, the span of E2, and the minimal sentential span encompassing E1 and E2 each serve as a separate input, allowing the model to explicitly address each of them as well as discover how these three inputs relate to one another.  This architecture is shown in Figure~\ref{fig:pitchfork-lstm-architecture}.  Each input feeds into its own LSTM and corresponding dropout layer before the three forks are merged via a concatenation of tensors.  Like the basic model, one version of the ``pitchfork'' is trained with vector weights initialized using the pre-trained word embeddings (FLSTM+P).  
% too soon
%The results of these experiments are shown in Table~\ref{tab:prec-results}.

% top-down chain for architecture
\begin{figure}[!htb]
\centering    
\scriptsize
\begin{tikzpicture}
  [trim left=-2.5cm,
  transform shape,
  node distance=.4cm,
  start chain=going below]
  % variables for positioning
  \def \xpos {2.7}
  \tikzset{
  % properties of all components
  component/.style={
    rectangle, 
    rounded corners,
    draw=black, very thick,
    text width=14em, 
    minimum height=1em, 
    text centered, 
    on chain
  },  
%  % layers
%    layer/.style={
%    fill=blue!10,
%    draw=blue, very thick,
%    text width=12em
%  },
  % output-layer
    output-layer/.style={
    fill=green!10,
    draw=green, very thick
  },
  % for brackets
  line/.style={draw, thick, <-},
  every join/.style={->, thick,shorten >=1pt},
  decoration={brace},
  % where to place text for bracket
  bracket/.style={decorate, midway, right=1pt},
}
% input
\node[component, join] (input-layer) {Input \\ (tokenized text)};
% hidden
\node[component, join] (embeddings-layer)      {Embeddings \\ (optional pre-trained weights)};
\node[component, join] (dropout)      {Dropout \\ (50\%)};
% output
\node[component, join] (dense)      {Dense \\ (dim. 3)};
\node[component, output-layer, join] (softmax)      {Softmax};
% output
% dedup bracket
\draw[bracket] let \p1=(input-layer.north), \p2=(embeddings-layer.south) in 
($(\xpos, \y1)$) -- ($(\xpos, \y2)$) node[bracket, align=left] {\hspace{.1cm} concatenated  vectors};
% ordering bracket
% dotted border
%\node[draw, dotted, no-foc, opacity=0.3, fit=(exact-match) (ml-dedup)] {};
%% ordering sieves
%\node[draw,dotted,fit=(intra-sentence) (io-order)] {};
\end{tikzpicture}
\caption{Architecture for the basic latent feature model using the minimal sentential span encompassing events 1 and 2 as input.}
\label{fig:lstm-architecture}
\end{figure}
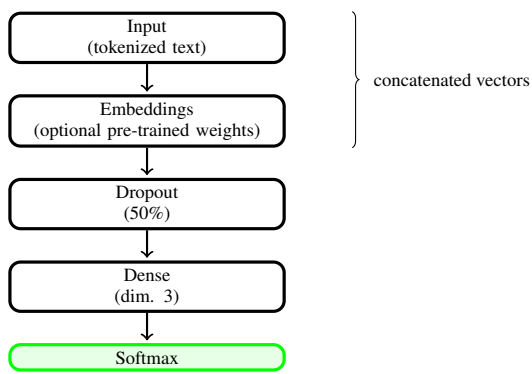

%pitchfork lstm architecture

%% file: results.tex
\section{Results}
We summarize the performance of all these models on the dataset previously introduced in Table~\ref{tab:prec-results}. 
We report results using micro precision, recall, and F1 scores for each model.
With fewer than 200 instances of causal precedence occurring in 1000 annotations, training and testing for both the feature-based classifiers and latent feature models was performed using stratified 10-fold cross validation.  
For the latent feature models, training was parameterized using a maximum of 100 epochs with support for early stopping through monitoring of validation loss\footnote{The validation set used for each fold came from a different class-balanced fold.}.  
Weight updates were made on batches of 32 examples and all folds completed in fewer than 50 epochs.

\begin{figure}[t]
\includegraphics[trim=0.25cm 0.25cm 1cm 0.5cm, clip, width=\columnwidth]{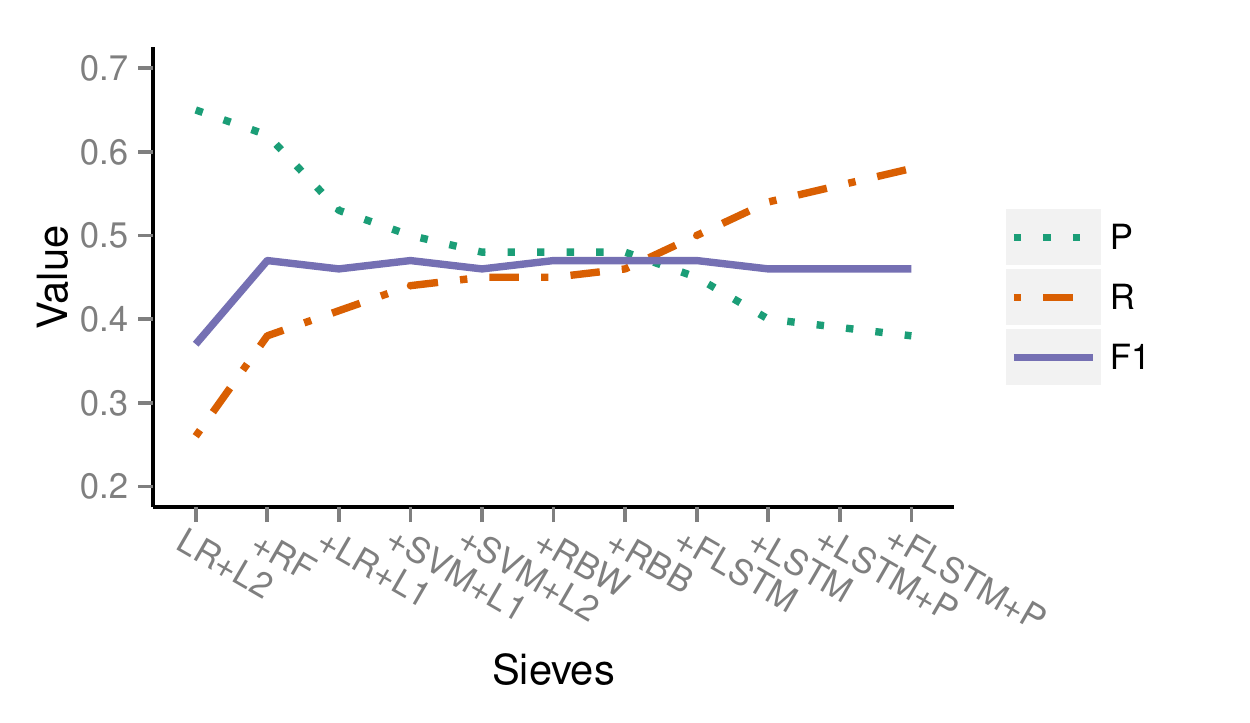}
\caption{The performance of the sieve-based combined model varies with each model added.}\label{fig:sieve performance}
\end{figure}

The table also includes a sieve-based ensemble system, which performs significantly better than the best-performing single model. In this architecture, the sieves are applied in descending order of precision, so that the positive predictions of the higher precision sieves will always be preferred to contradictory predictions made by subsequent, lower-precision sieves. Figure \ref{fig:sieve performance} illustrates that as sieves are added, the F1 score remains fairly constant, while recall increases at the cost of precision.

%so that the positive predictions of the higher precision sieves will always be preferred to contradictory predictions made by subsequent, lower-precision sieves.
% combined recall = (43 + 47 + 34) / 194 =  0.64
% averaged micro precision for models = ((58 + 65 + 54 + 54 + 62) + (40 + 39 + 43 + 38) + (80 + 50)) / 11 = 53
% unreliable F1 estimate: (64 * 2 * 53) / (64 + 53) = 58

% TODO: Do bootstrap significance tests

\begin{table}[!htb]
%\small
\centering
\begin{tabular}{l l l l}
	\toprule
    \textit{Model} & \textit{p} & \textit{r} & \textit{f1} \\
    	\midrule
    Intra-sentence & 0.5 & 0.01 & 0.01 \\
    Inter-sentence & 0.5 & 0.01 & 0.01 \\
    Reichenbach & 0 & 0 & 0 \\
    \midrule
    LR+L1 	& 0.58	& 0.32	& 0.41 			\\
    LR+L2 	& 0.65	& 0.26	& 0.37 			\\
    SVM+L1 	& 0.54	& 0.35	& \textbf{0.43}	\\
    SVM+L2 	& 0.54	& 0.29	& 0.38 			\\
    RF 		& 0.62	& 0.25	& 0.36 			\\
    \midrule
    LSTM     & 0.40& 0.25 & 0.31 \\
    LSTM+P   & 0.39 & 0.20 & 0.26 \\
    FLSTM    & 0.43 & 0.15 & 0.22 \\
    FLSTM+P  & 0.38 & 0.22 & 0.28 \\
    \midrule
    Combined & 0.38 & 0.58 & \textbf{0.46*} \\
    \bottomrule
\end{tabular}
    
\caption{Results of all proposed causal models, using stratified 10-fold cross-validation. The combined system is a sieve-based architecture that applies the models in decreasing order of their precision. The combined system significantly outperforms the best single model, SVM with L1 regularization, according to a bootstrap resampling test (p = 0.022).}
\label{tab:prec-results}
\end{table}

%\begin{table}[h]
%\small
%\centering
%\begin{tabular}{r p{1.75cm} l@{\hskip 0.5em} l@{\hskip 2em} l@{\hskip 0.5em} l@{\hskip 2em} l@{\hskip 0.5em}l}
%	\toprule
%    \multicolumn{2}{c}{\textit{Features}} & \textit{p} & \% & \textit{r} & \% & \textit{f1} \% \\
%    	\midrule
%    \multicolumn{2}{l}{All features} & 0.54 &  & 0.35 & & 0.43 &	\\[0.3em]
%    $-$ & Event	& 0.47 & -?? & 0.34 & -?? & 0.39 & -?? \\
%    $-$ & {Event-Event \newline (surface)} & 0.5 & ?? & 0.32 & ?? & 0.39 & ?? \\
%    %0.52	0.32	0.4
%    $-$ & {Event-Event \newline (syntax)} & ?? & ?? & ?? & & ?? & ??	\\
%    $-$ & Coreference & ?? & ?? & ?? & ?? & ?? & ??	\\
%    \bottomrule
%\end{tabular}
%\label{tab:ablation-results}
%\caption{\todo{Describe ablation results}}
%\end{table}

Despite some obvious patterns noted in Table~\ref{tab:label-examples}, the deterministic models perform the worst due in large part to their rarity in the corpus.  An analysis of this result is given in Section~\ref{sec:discussion}. 
Overall, our top-performing model was the linear kernel SVM with L1 regularization.  
%The effectiveness of L1 regularization here is likely due to its aggressive discount of features.  
In all cases, the feature-based classifiers outperform the latent feature representations, suggesting that in cases such as this where little data is available, feature-based classifiers capitalizing on high-level linguistic features are able to better generalize and thus outperform latent feature models. However, as our discussion in Section~\ref{sec:overlap} will show, our combined model demonstrates that the latent and feature-based models are largely complementary.

%% file: discussion.tex
\section{Discussion}
\label{sec:discussion}

Overall, results are promising, particularly in light of the conscious choice to omit (causal) regulation reactions from this task, as they are already captured by the Reach reading system.

However, the deterministic models created so far have extremely low recall, such that 
it is difficult even to determine their precision. An analysis of the 
Reichenbach model reveals one source of this low coverage. In short, 
although writers \textit{could} describe causal mechanisms using temporal 
indicators such as tense and aspect, temporal description is rare enough in 
this domain not to be represented in our randomly sampled database. Table 
\ref{tab:reichenbach} illustrates the lack of overlap with informative tense-aspect combinations; a single tense is used per passage, and no
perfective aspect is used.

\begin{table}[!htb]
\scriptsize
\centering
\begin{tabular}{l@{\hskip 1em}l c c c c c c}
\toprule
\multicolumn{2}{c}{\multirow{2}{*}{E1$\downarrow$, E2$\rightarrow$}}	& 
\multicolumn{2}{c}{\textit{past}}	& 
\multicolumn{2}{c}{\textit{pres.}}	& 
\multicolumn{2}{c}{\textit{fut.}}	\\
					& 	& 
					\textit{simple}	& 
					\textit{perf.}	& 
					\textit{simple}	& 
					\textit{perf.}	& 
					\textit{simple}	& 
					\textit{perf.}	\\
\midrule
\multirow{2}{*}{\textit{past}}		& \textit{simple}	& 
69	& \cellcolor{tbl}0	& 38		& 0	& \cellcolor{tbl}0	& \cellcolor{tbl}0	\\
							& \textit{perf.}	& \cellcolor{tbl}0	& 0	& \cellcolor{tbl}0		& \cellcolor{tbl}0	& \cellcolor{tbl}0	& \cellcolor{tbl}0	\\
\midrule
\multirow{2}{*}{\textit{pres.}}		& \textit{simple}	& 49	& \cellcolor{tbl}0	& 134	& 0	& \cellcolor{tbl}1	& 0	\\
							& \textit{perf.}	& 0	& \cellcolor{tbl}0	& 0		& 0	& \cellcolor{tbl}0	& \cellcolor{tbl}0	\\
\midrule
\multirow{2}{*}{\textit{fut.}}		& \textit{simple}	& \cellcolor{tbl}0	& \cellcolor{tbl}0	& \cellcolor{tbl}0		& \cellcolor{tbl}0	& 0	& 0	\\
							& \textit{perf.}	& \cellcolor{tbl}0	& \cellcolor{tbl}0	& 0		& \cellcolor{tbl}0	& 0	& 0	\\
\bottomrule
\end{tabular}
\caption{Event tense and aspect for events containing verbs in the present
study. Highlighted cells are tense-aspect combinations that are informative 
for establishing temporal precedence, following \newcite{chambers2014dense}. 
All but one event pair fall outside of these informative combinations, and that exceptional pair was a false positive case.}\label{tab:reichenbach}

\end{table}

Similarly, the time expressions required by the deterministic intra- and 
inter-sentence precedence rules are rare enough to make them ineffective on 
this sample.

\subsection{Model overlap}
\label{sec:overlap}

\begin{figure}[!htb]
\centering
\begin{subfigure}{\columnwidth}
	\centering
	\includegraphics[trim=0 1.5cm 0 1.5cm, clip, width=0.75\columnwidth]{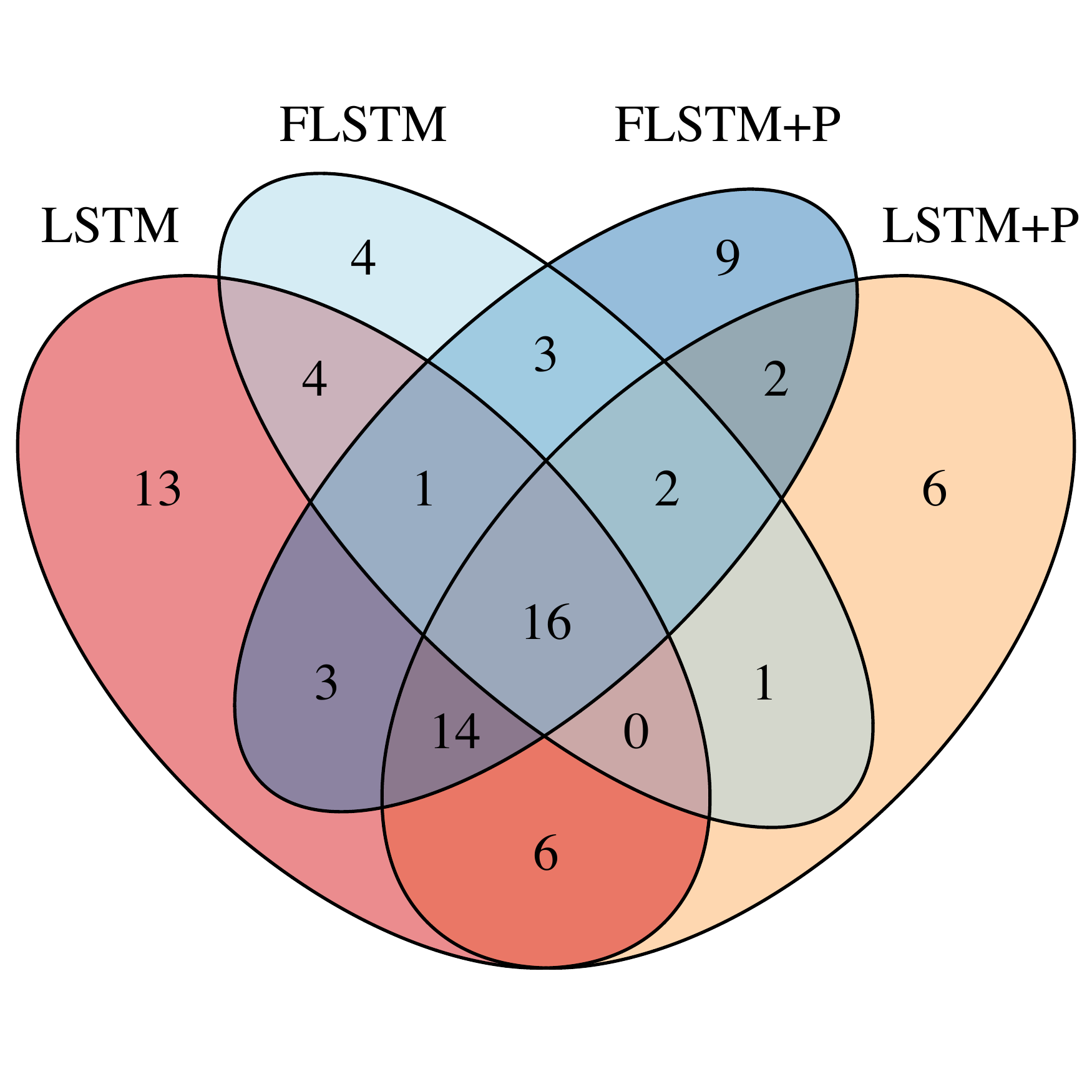}
	\caption{Overlap of true positive predictions made by LSTM models.  Though in Table~\ref{tab:prec-results} the models appear to perform similarly, the learned representations are largely distinct and complementary in their coverage.}
\end{subfigure}\\

\begin{subfigure}{\columnwidth}
	\centering
	\includegraphics[trim=4cm 5cm 3.5cm 3cm, clip, width=0.75\columnwidth]{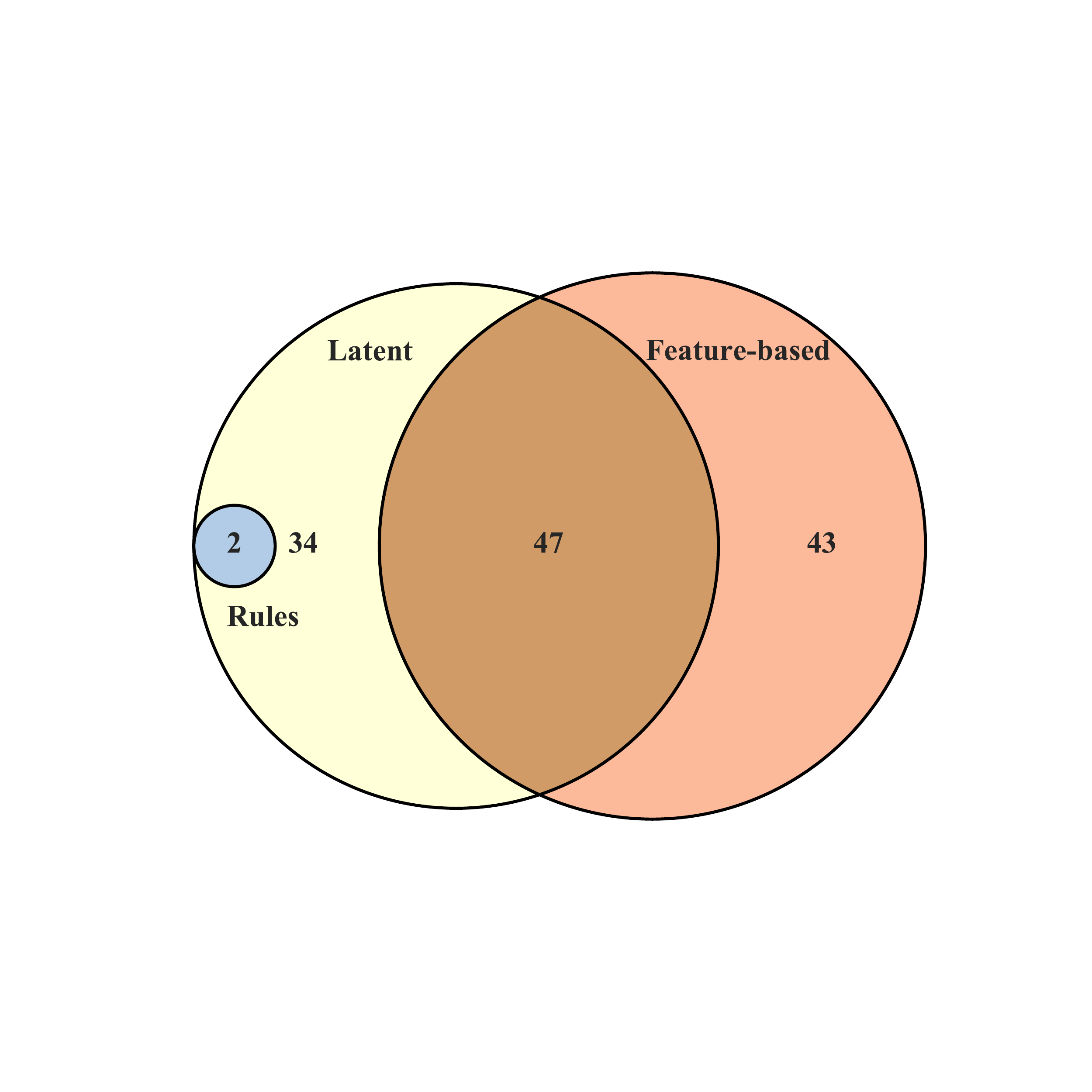}
	\caption{Similarly, the overlap between the feature-based models and the latent models was low overall.}
\end{subfigure}
\caption{The overlap of true positives among the investigated models was low.}\label{fig:venn}
\vspace{-4mm}
\end{figure}

As \newcite{chambers2014dense}, \newcite{1604.08120}, and many other algorithms have shown, models can be applied sequentially in ``sieves'' to produce higher-quality output.  Ideally, each model in a sieve-based system will capture different portions of the data through a mixture of approaches, distinguishing this method from more naive ensembles in which the contributions of a lone component would be washed out. 
%Our combined system suggests that the models investigated here are fit for an efficient ensemble setup such as a sieve architecture.
Figure \ref{fig:venn} details this observation by showing the coverage difference between the models described here.

\subsection{Error analysis}
We performed an analysis of the false positives shared by all feature-based classifiers, in addition to the false negatives shared by all models. Here we limit our discussion to only the most prominent characteristic shared by the majority of false positives.

\paragraph{Discourse information}

More than half of the false positives share contrastive discourse features, suggesting that a model of discourse could improve classifier discrimination. Example (\ref{ex:discourse}) demonstrates such a contrastive structure, which \textit{whereas} introduces a clause (and event) that is contrasted and therefore both temporally and causally distinct from the following clause (and event). The existence of regular cues like \textit{whereas} indicates that a feature to explicitly model these structures is possible.

\begin{exe}
	\ex\label{ex:discourse} Whereas $[$PRAS40 inhibits the mTORC1 activity via raptor$]$\textsubscript{E1}, DEPTOR was identified to interact directly with mTOR in both $[$mTORC1 and mTORC2 complexes$]$\textsubscript{E2} 
%\textsubscript{after}
\end{exe}

%% file: related.tex
\section{Related Work}

Though focused on temporal ordering, \newcite{chambers2014dense} adopt a sieve-based approach, with high-precision deterministic sieves preceding and constraining lower-precision, higher-recall machine learning sieves. As with our system, the deterministic sieves were linguistically motivated, and had the additional advantage of operating over time expressions (\textit{during}, \textit{Friday}, etc.) as well as events, the former of which are typically lacking in the biomedical domain.

\newcite{1604.08120} implemented a hybrid sieve-based approach for causal relation detection between events that includes a set of causal verb rules and corresponding syntactic dependencies and a feature-based classifier. 
However, both of these works focus on open-domain texts. To our knowledge, we are the first to investigate causal precedence in the biomedical domain.
%"We combine the rule-base methods presented in Mirza et al. (2014) with the statistical-based methods presented in Mirza and Tonelli (2014a) in a similar fashion as for temporal relation extraction."